# A New Automatic Method to Adjust Parameters for Object Recognition

Issam Qaffou

Département Informatique, FSSM
Université Cadi Ayyad,
Marrakech, Morocco

Mohamed Sadgal

Département Informatique, FSSM
Université Cadi Ayyad
Marrakech, Morocco

Aziz Elfazziki

Département Informatique, FSSM
Université Cadi Ayyad
Marrakech, Morocco

*Abstract—* **To recognize an object in an image, the user must apply a combination of operators, where each operator has a set of parameters. These parameters must be "well" adjusted in order to reach good results. Usually, this adjustment is made manually by the user. In this paper we propose a new method to automate the process of parameter adjustment for an object recognition task. Our method is based on reinforcement learning, we use two types of agents: User Agent that gives the necessary information and Parameter Agent that adjusts the parameters of each operator. Due to the nature of reinforcement learning the results do not depend only on the system characteristics but also the user's favorite choices.**

## I. Introduction

New tools and new algorithms for vision applications cause new system parameters that must be properly adjusted. This adjustment requires a specific knowledge, takes a long time, and sometimes even has to be done in an experimental process. To accomplish a segmentation task, the user must apply some operators, where each one has a set of parameter to adjust. The lack of a general rule that guides the user in his choices, the fixation of parameter values is usually made intuitively. The user proceeds by trying manually all possible cases until finding the desired result. Usually, in the majority of vision tasks we need to apply a combination of several operators where each one has a multitude of parameters to adjust. So, the manual adjustment becomes very tedious and not trustworthy. Therefore, an automatic method to adjust the values of each parameter is needed. The quality of results depends essentially on the operator chosen and the values assigned to its parameters.

Some GUI, for example Ariane [1], help users to accomplish a vision task by proposing them an interactive interface, but the values assigned to the parameters are selected manually by the user. Very few systems had succeeded to automate the process of parameter adjustment.

In [2], B.NICKOLAY et al. proposed a method to automatically optimize the parameters of a machine vision system for surface inspection by using specific Evolutionary Algorithms (EA). A few years later, Taylor proposed a reinforcement learning framework which uses connectionist systems as function approximators to handle the problem of determining the optimal parameters for a computer vision application even in the case of a highly dimensional, continuous parameter space [3]. More recently, Farhang et al. [9] introduced a new method for the segmentation of the prostate in transrectal ultrasound images, using a reinforcement learning (RL) scheme. He divided the initial image into sub-images and works on each one in order to reach a good result.

In this paper we propose a new method to adjust automatically the parameters of vision operators. Our method is based on reinforcement learning. We use two agents: User Agent (UA) and Parameter Agent (PA). The UA gives the necessary information to the system. It gives the combination of applicable operators, the set of adjustable parameters for each operator, values' ranges for each parameter. The PA uses reinforcement learning to assign the optimal values for each parameter in order to extract the object of interest from an image.

Due to the nature of RL, in terms of the interaction between state, action and reward, our approach takes in account not only the system opportunities but also the user preferences, and through the learning mechanism it will suggest trustworthy solutions.

An overview of reinforcement learning is given in section 2. Section 3 outlines the proposed approach and introduces a general framework for parameter adjustment. Section 4 presents the experimental results, and section 5 concludes the paper.

## II. Reinforcement Learning

Reinforcement learning (RL) is learning what to do, how to map situations to actions, so as to maximize a numerical reward signal. The learner is not told which actions to take, as in most forms of machine learning, but instead must discover which actions yield the best reward by trying them. One of the challenges that arise in reinforcement learning and not in other kinds of learning is the tradeoff between exploration and exploitation. To obtain a lot of reward, a reinforcement learning agent must prefer actions that it has tried in the past and found to be effective in producing reward. But to discover such actions it has to try those that it has not selected before.

Reinforcement learning uses a formal framework defining the interaction between agent and its environment in terms of states, actions, and rewards, Fig 1.

Reward or punishment is determined from the environment, depending on the action taken. The agent must find a trade-off between immediate and long-term returns. It must explore the unseen states, as well as the states which maximize the return by choosing what the agent already knows. Therefore, a balance between the exploration of unseen states and the exploitation of familiar (rewarding) states is crucial. Watkins has developed Q-learning, a well-established on-line learning algorithm, as a practical RL method [6]. In this algorithm, the agent maintains a numerical value for each state-action, representing a prediction of the worthiness of taking an action in a state.

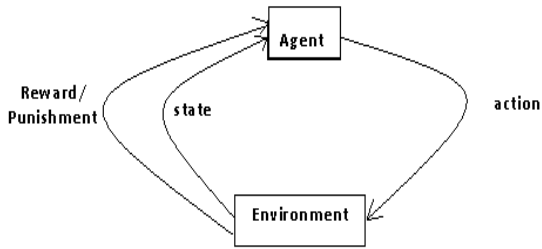

Figure 1: A general model for Reinforcement learning agent

Table 1 represents an iterative policy evaluation for updating the state-action values where r is the reward value received for taking action in states, s' is the next state, α is the learning rate, and γ is the discount factor [7]. There are some policies for taking action a given states. One of them is the Boltzman policy which estimates the probability of taking each action in each state. There are other policies for Q-learning such as ε- greedy and greedy. In the greedy policy, all actions may not be explored, whereas the ε-greedy selects the action with the highest Q-value in the given state with a probability of 1 − ε, and other ones with a probability of ε [7,8]. In this work an ε-greedy policy is used to make a balance between exploration and exploitation. The reward r(s, a) is defined according to each state-action pair (s, a). The goal is to find a policy to maximize the discounted sum of rewards received over time. The principal concerns in RL are the cases where the optimal solutions cannot be found, but can be approximated. The online nature of RL distinguishes it from other techniques that approximately solve Markov decision processes (MDP) [5,7].

TABLE 1. Q-LEARNING ALGORITHM

| |
|---|
| Initialize $Q(s, a)$ arbitrary |
|   Repeat (for each episode) |
|    Initialize state s |
|    Repeat (for each step of episode) |
|    Choose action $a$ from state s using policy derived from $Q$ (e.g., $\varepsilon - greedy$) |
|    Take action $a$, observe reward r, next state s' |
|      $Q(s, a) \leftarrow Q(s, a) + \alpha[r + \gamma \max_{a'} Q(s', a') - Q(s, a)]$ |
|      $s \leftarrow s';$ |
|    Until s is terminal |

In this paper, we attempt to introduce the RL concept for parameters adjustment.

## III. THE PROPOSED APPROACH

Generally, to accomplish an object recognition task, the user must apply sequentially some operators, and for each operator there is some parameter to adjust. Because there is no general rule that guides the user in his choices, he is based usually on his intuition to select values for each parameter. In the majority of vision tasks, we have to apply a multitude of operators that have several parameters to adjust. So adjusting manually these parameters basing only on the experience and on the intuition is not evident. It's a tedious work with a huge wasted time. In this paper we propose a new automatic method to find the best values for each parameter in a recognition task.

In our method we use two types of agents: User Agent (UA) and Parameter Agent (PA). Fig 2 shows the general framework of our method.

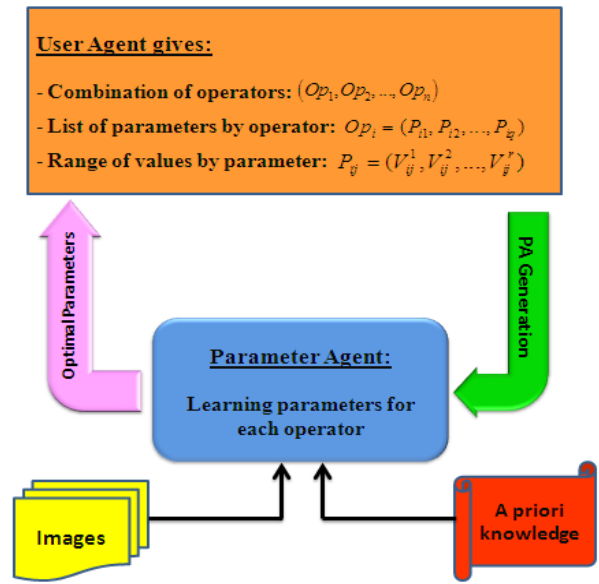

Figure 2: General framework for the proposed approach.

The UA gives to the PA the needed information: the combination of operators to apply, the set of parameters for each operator and the values' ranges for each parameter.

The PA receives this information and proceeds automatically to find the best values for each parameter. Fig. 3 shows the general functioning of the PA.

The agent PA interacts with its environment by actions, states. A set of images containing the object of interest is given to PA. Each image has its ground-truth, the object extracted by an expert.

An image with its ground truth is introduced to the system. A combination of operators to extract an object of interest is proposed. Each operator has some parameters that have to be well adjusted. Each value given to a parameter gives a different result. The agent PA must find the optimal values that give the best result. It proceeds then by trial and error until finding the best parameter values. For that it uses reinforcement learning. Actions, states and a reward function must then be defined.

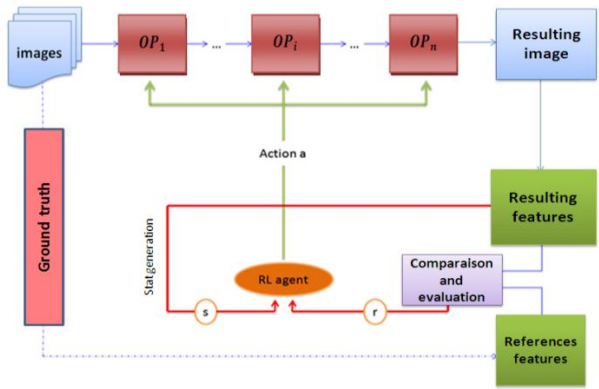

Figure 3: the general process of PA using reinforcement learning.

### A. *Defining actions*

Generally, all possible combination of parameters values is defined as an action for the RL agent. The set of the actions is then the set of all possible values combination, see fig. 3.

Each operator $OP_k$ has a series of parameters:

$$(P_1^k, P_2^k, ..., P_n^k)$$

Each parameter $P_j^k$ has a range of values:

$$V_j^k = \{V_{j1}^k, V_{j2}^k, ..., V_{jm}^k\}$$

An elementary action of the operator $OP_k$ is:

$$a_k = (u_{j1}^k, ..., u_{jr}^k) \text{ where } u_{j1}^k \in V_j^k$$

An action of the agent PA is defined by the combinations of the elementary actions of operators as it is defined above:

$$a = (a_1, a_2, ..., a_n)$$

Actions for object recognition task are given in the experience.

### B. *Defining states*

A state is defined by a set of features extracted from the resulting image:

$$s = [\chi_1, \chi_2, ..., \chi_n]$$

$\chi_i$ is a feature reflecting the state of the image after the processing. The type of the extracted features depends on the task at hand. Here we give a general definition, and in the experience we define them explicitly for a recognition task.

### C. *Defining the reward*

The return is a reward if the agent chooses the right action, else it is a punishment. The reward is defined according to the quality of the processing result. This quality is assessed by using ground-truth models. To define the return we calculate the similarity between the resulting image and its ground truth. The similarity is calculated according to some features extracted from the two images. The type of these features depends on the task at hand. For example, if we want to detect an object in an image we extract the number of the objects, their areas, their sizes, etc. We express the difference between these scalars by:

$$D = \sum_i w_i D_i$$

The weights $w_i$ are chosen according to the importance of each feature.

A general form of the reward definition in the proposed approach is presented by:

Reward: r= -10, 0 or 10;
    if (D < ε) r = +10; f=true;
        elseif ( (D > ε) && (D < ε + δ) )
          r = 0;
        else r = -10;
    end
    end

The values 10 and -10 represent respectively the reward and the punishment depending to a predefined threshold.

## IV. EXPERIENCE

We use a dataset of 30 textured images containing the same object to extract. The object is a textured disc injected in all the 30 images. The used images are textured so the UA proposes a combination of two operators: GLCM (Gray Level Cooccurrence Matrices) to segment textures and k-means to classify them. Each one of these operators has some parameters to be adjusted in order to be executable. In GLCM, texture is always defined in relation to some local window. The size n x n of this window affects the result of the segmentation, so we propose the size of the window as the parameter to adjust for GLCM. UA proposes a range of values for n, it may have seven values, the odd values between 9 and 21 {9, 11,..., 21}. GLCM extracts fourteen texture features [8]. In this paper we limit our self to four of the most popular features: Angular Second Moment (energy), Contrast, Correlation and entropy. After extracting these textures, we classify them using the algorithm k-means. The parameter to adjust for this operator is k, the number of the possible clusters. It can take five values {1,...,5}. How are actions, states and reward are defined according to our experience is given below.

### A. *Actions Definition*

The UA proposes two operators: GLCM and k-means. GLCM has n the size of the sliding window as the parameter to adjust. n can take seven odd values: 9, 11, 13, ..., 21, so an elementary action for GLCM is one of these values. k-means has k the number of possible cluster, its possible values are {1,2,3,4,5}. An elementary action for k-means is one of these values.

Then an action for the agent PA is constituted by a couple of a value of "n" and a value of "k". All actions are all possible combinations of the values of "n" and "k".

### B. *States Definition*

States are defined, according to the features which represent the status of the resulting image. For object recognition we extract four features to define the state space:

$$S = [x_1, x_2, x_3, x_4] \qquad (1)$$

Where $x_i$ is the selected feature.

$x_1$ is the Number of Objects in the resulting image after segmentation.

$x_2$ is the ratio between the area of the extracted object and the area of the whole image.

$x_3$ is the ratio between the area of the resulting object and the object reference.

$x_4$ is the mean of the used textural features: Angular Second Moment (energy), Contrast, Correlation and entropy.

*C. Reward Definition*

The rewards and punishments are defined according to the quality criterion that represents how well the image is segmented. A straightforward method is comparing the resulting image with its ground truth. This comparison is made between the scalar features of the obtained regions and those of the desired one. In this paper we define the reward according to a difference between the components of the image. We define this difference as: D = weighted sum of the four following differences in the two images (the resulting image and its ground truth):

D1= difference of the number of the objects;

D2= difference of the sizes of the objects;

D3= difference of the surfaces of the objects;

D4= difference of the feature textures.

$$D = \sum_i w_i D_i$$

Fig. 4 shows the three images taken randomly from the process of recognition. The three images contain the same object of interest, the textured disc.

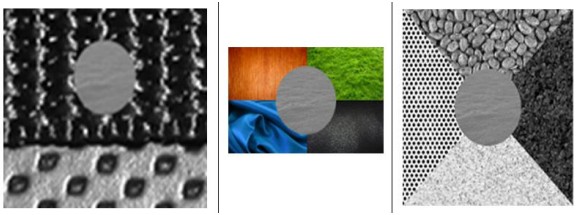

Figure 4: the three images containing the disc to extract.

The agent PA proceeds by reinforcement learning and finds that the optimal action that gives the best result is $(t_w = 13, Nb_c = 3)$. So the best value for the size of the silding window is 13 with the best number of possible clusters is 3. Fig. 5 shows the reference disc and the resulting one by our approach.

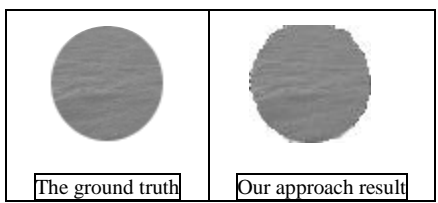

Figure 5: the resulting image and its reference.

Fig. 6 shows the curve of learning of the agent PA. At first it has not much knowledge and experience to behave, so it uses several steps per episode.

Over time the curve learning becomes almost constant, which proves that really there is a learning while the processing is done, the number of steps decreases with episodes. It means that our agent RL accumulates an experience that will help him to take decision in the future.

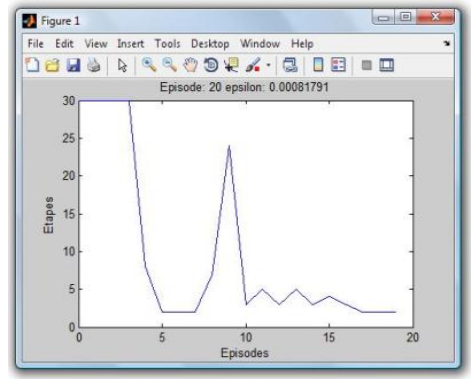

Figure 6: Learning that makes our RL agent during its processing

V. CONCLUSION

Determining the values of parameters of the vision operators is a challenging task. In this paper, we have proposed a reinforcement learning approach to handle this problem even in the case of a vision task needing many operators to sequence. A texture segmentation application is presented to test our approach.

Our goal isn't comparing our method to others, but our goal is to present another manner of thinking that uses learning concepts and show that really it gives good results. Our method can be applied to any decision process using parametric methods.

Due to the nature of reinforcement learning, the proposed approach takes in account not only the system opportunities but also the user preferences, and through the learning mechanism it suggests trustworthy solutions. As perspectives, our approach will be used on a large set of different images and its results will be compared to other methods.